\documentclass[cameraready]{Interspeech}

\title{HATS: An Open data set Integrating Human Perception Applied to the Evaluation of Automatic Speech Recognition Metrics}

\author[affiliation={1}, orcid=0000-0000-0000-0000]{Thibault}{Ba\~{n}eras Roux}
\author[affiliation={2}, orcid=0000-0000-0000-1111]{Jane}{Wottawa}
\author[affiliation={3}, orcid=0000-0000-0000-1111]{Mickael}{Rouvier}
\author[affiliation={2}, orcid=0000-0000-0000-1111]{Teva}{Merlin}
\author[affiliation={1}, orcid=0000-0000-0000-1111, correspondingauthor]{Richard}{Dufour}

\address{
    $^1$ Nantes University, France \\
    $^2$ Le Mans University, France \\
    $^3$ Avignon University, France
}

\email{\{thibault.roux,richard.dufour\}@univ-nantes.fr, jane.wottawa@univ-lemans.fr, \{mickael.rouvier,teva.merlin\}@univ-avignon.fr}


\usepackage{comment}

\usepackage{multirow}

\begin{document}

\maketitle

\begin{abstract}
    Conventionally, Automatic Speech Recognition (ASR) systems are evaluated on their ability to correctly recognize each word contained in a speech signal. In this context, the word error rate (WER) metric is the reference for evaluating speech transcripts. Several studies have shown that this measure is too limited to correctly evaluate an ASR system, which has led to the proposal of other variants of metrics (weighted WER, BERTscore, semantic distance, etc.). However, they remain system-oriented, even when transcripts are intended for humans. In this paper, we firstly present \textbf{H}uman \textbf{A}ssessed \textbf{T}ranscription \textbf{S}ide-by-side (HATS), an original French manually annotated data set in terms of human perception of transcription errors produced by various ASR systems. 143 humans were asked to choose the best automatic transcription out of two hypotheses. We investigated the relationship between human preferences and various ASR evaluation metrics, including lexical and embedding-based ones, the latter being those that correlate supposedly the most with human perception. 
\end{abstract}

\section{Introduction}
Automatic Speech Recognition (ASR) consists in transcribing speech into its textual form. Automatic transcriptions can for example be used by humans in the case of captioning, speech-to-text messages or by third systems such as virtual personal assistants. Since the emergence of hidden Markov model-based ASR systems~\cite{juang1991hidden} for processing continuous speech, the field has seen an important breakthrough with the use of deep neural networks and self-supervised methods such as wav2vec~\cite{baevski2020wav2vec} and HuBERT~\cite{hsu2021hubert}. These approaches allow the extraction of meaningful information from speech without previously labeled data.

Faced with transcription errors, unlike a machine, a human is able to process the sentence anyway and extract its initial meaning if the latter was not fundamentally impacted by the errors. 
Errors in automatic transcriptions can arise due to various factors such as noise in the speech signal, speaker accents, or technical limitations. The question is to determine which errors are acceptable and which ones may cause comprehension difficulties for humans. 
Thus, it is crucial to evaluate the quality of automatic transcriptions based on their overall comprehensibility to humans. 

Currently, the most commonly used metrics for evaluating ASR systems are the Word Error Rate (WER), which measures the number of incorrectly transcribed words, and the Character Error Rate (CER), which calculates the number of characters that differ from the reference transcription. However, many researchers~\cite{wang2003word,favre2013automatic,itoh2015metric,kafle2017evaluating} have pointed out issues with these metrics, such as the absence of error weighting or the lack of linguistic and semantic knowledge.
Consequently, there has been a growing interest in developing new metrics to evaluate ASR systems.  
Some researchers~\cite{le16_interspeech,nam2019simulation,gordeeva2021meaning,kim2021semantic,roux2022qualitative} have therefore started exploring alternative metrics that can more accurately assess the quality and effectiveness of automatic transcriptions. Similarly, these issues have been observed in the field of machine translation. As a result, new metrics and data sets have been produced from multiple shared tasks~\cite{mdhaffar2019qualitative,mathur2020results,freitag2021results,freitag2022results}. Semantic-based metrics, such as BERTScore~\cite{zhang2019bertscore}, have then been shown to be effective in evaluating the quality of machine-generated translations.

\begin{figure}[thb]
\centering
\includegraphics[scale=0.35]{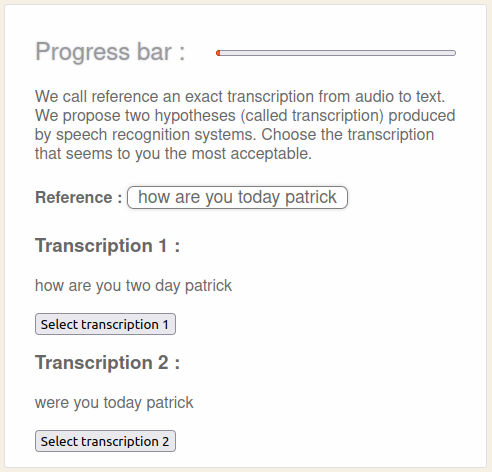}
\caption{Screenshot from the side-by-side experiment.}
\label{fig:perceptualexp}
\end{figure}

While these metrics are obtained automatically and are rather {\it machine-oriented}, human evaluations of ASR systems have been carried out in the past, which includes side-by-side experiments~\cite{gordeeva2021meaning,kafle2017evaluating,kim2021evaluating} where human subjects are asked to choose the best transcript among two options. These studies have also enabled assessing the quality of automatic metrics from a human perspective.
The present study builds on these side-by-side experimental protocols, but instead of modifying the speech signal or text hypothesis with artificially generated errors, or using different outputs from the same ASR system to obtain two different hypotheses, our study utilizes the outputs of ten ASR systems with varying architectures applied on the same speech corpus. Furthermore, rigorous criteria were used to select the transcripts where choices are the harder in order to study metric and human behavior. 
The advantage of the side-by-side experiment is that the subject has to make a choice between two hypotheses, which does not allow for equality. In contrast to direct assessment, side-by-side experiments eliminate the potential bias of prior choices, allowing for consistent comparisons between transcriptions. By comparing human judgments to those of the metric, we can effectively evaluate its performance.

In this paper, we introduce HATS (Human-Assessed Transcription Side-by-Side), a new open data set of human preferences on erroneous transcriptions in French from various ASR architectures. As a second contribution, an original study is conducted using HATS to evaluate automatic metrics by analyzing their agreement with human assessments. Our objective is to identify the ASR evaluation metrics that most closely correlate with human perception. The HATS data set is freely released to the scientific community\footnote{\url{https://github.com/thibault-roux/metric-evaluator}}.


The paper is organized as follows: Section~\ref{sec:systems-metrics} describes the used ASR systems and the automatic metrics that will be evaluated based on their correlation with human perception. In Section~\ref{sec:human_eval}, we present the implementation of the side-by-side human perception experiment, including the protocol for selecting the transcripts provided to human evaluators. Section~\ref{sec:HATS_set} describes the HATS data set, while Section~\ref{sec:metric_evaluation} presents a study on the quality of automatic metrics for evaluating transcription systems in relation to human perception. Finally, Section~\ref{sec:conclusion} provides the conclusion and future work.

\section{Transcription systems and ASR evaluation metrics}
\label{sec:systems-metrics}

In Section~\ref{s:asr}, we present the different automatic speech recognition systems used to obtain the automatic transcriptions that constitute the HATS corpus. Then, in Section~\ref{sec:metrics}, we describe all the evaluation metrics applied to assess these transcriptions and evaluate them in relation to human perception.

\subsection{Automatic transcription systems}
\label{s:asr}


In this study, we set up 8 end-to-end systems based on the Speechbrain toolkit~\cite{ravanelli2021speechbrain} and 2 DNN-HMM-based systems using a state-of-the-art recipe\footnote{\url{https://github.com/kaldi-asr/kaldi/blob/master/egs/librispeech/s5/}} with the Kaldi toolkit~\cite{povey2011kaldi}. 
The end-to-end ASR systems were trained using various self-supervised acoustic models. Seven of the systems used variants of the wav2vec2 models learned on French~\cite{evain2021task}, and one system used the XLS-R-300m model. In the Kaldi pipeline systems, one of the systems included an extra rescoring step using a neural language model. 


All ASR systems have been trained to process French using ESTER 1 and 2~\cite{galliano2006corpus,galliano2009ester}, EPAC~\cite{esteve2010epac}, ETAPE~\cite{gravier2012etape}, REPERE~\cite{giraudel2012repere} train corpora, as well as internal data. Taken together, the corpora represent approximately 940 hours of audio comprised of radio and television broadcast data. 
The transcripts used to build our HATS corpus are extracted from the REPERE test set, which represents about 10 hours of audio data.

\subsection{Evaluation metrics}
\label{sec:metrics}

We propose to focus on evaluation metrics for transcription systems that enable us to evaluate the systems at both lexical and semantic levels. First of all, we consider classical lexical metrics such as {\bf Word Error Rate} and {\bf Character Error Rate}.

Next, we examine three semantic metrics based on word embedding representations.  
The first one, {\bf Embedding Error Rate (EmbER)}~\cite{roux2022qualitative}, is a WER where substitution errors are weighted according to the cosine distance between the reference and the substitute word embeddings obtained from fastText~\cite{grave2018learning,bojanowski2017enriching}. The second one, \textbf{SemDist}~\cite{kim2021semantic}, involves calculating the cosine similarity between the reference and hypothesis using embeddings obtained at the sentence level. 
We compared different pre-trained word embedding models to evaluate their impact on the metric. Specifically, we compared using the embedding of the first token from CamemBERT~\cite{martin2020camembert} or FlauBERT~\cite{le2020flaubert} models, or using the output of a sentence embedding model (SentenceBERT~\cite{reimers2019sentence}).
Our last semantic metric is \textbf{BERTScore}~\cite{zhang2019bertscore}, that computes a similarity score for each token in the candidate sentence with each token in the reference sentence using contextual embeddings. In our study, we use a multilingual BERT~\cite{devlin2018bert} and CamemBERT\footnote{\url{https://camembert-model.fr}}~\cite{martin2020camembert} models (both CamemBERT-base and CamemBERT-large). 

\begin{figure}[htb!]
    \centering
   \includegraphics[width=0.45\textwidth]{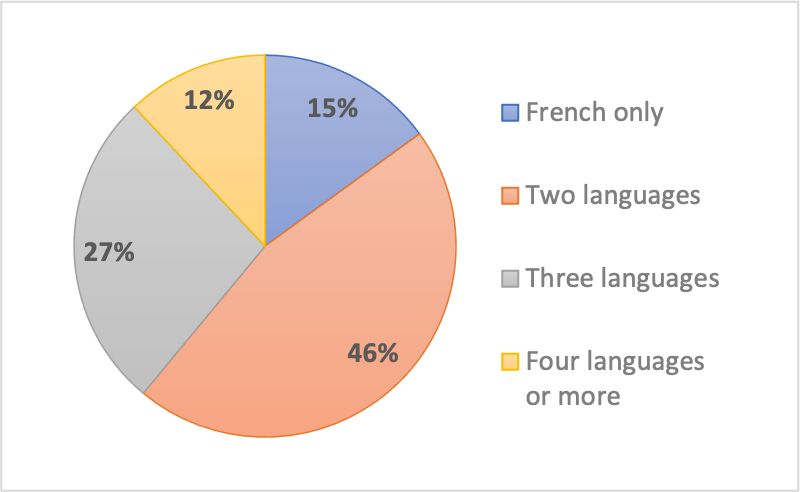}
    \caption{Participant characterization in terms of number of spoken languages.}
    \label{fig:spk_characterization}
\end{figure}

\begin{figure}[htb!]
    \centering
    \includegraphics[width=0.45\textwidth]{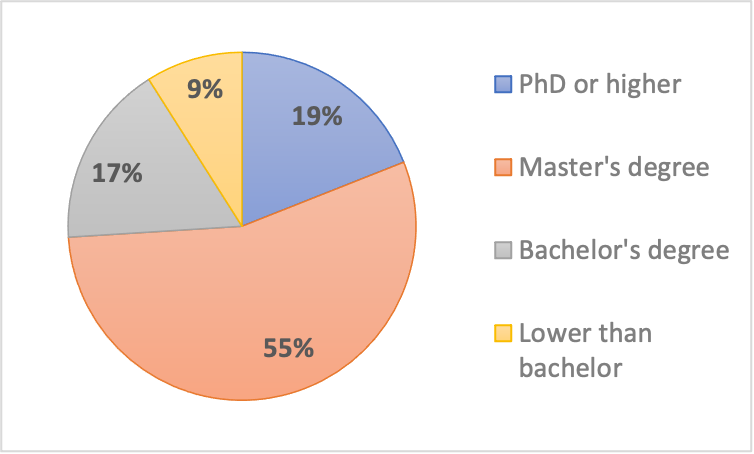}
    \caption{Participant characterization in terms of level of education.}
    \label{fig:spk_education}
\end{figure}

\begin{table*}[htp]
    \centering
    \caption{Detail of some stimuli choice criteria with examples. The $\epsilon$ symbol represents a missing word.}
    \scalebox{0.81}{
        \begin{tabular}{ccccc}
        \hline
        \multirow{2}{*}{\textbf{Category}} &  \textbf{Metrics} & \multirow{2}{*}{\textbf{Reference}} & \multirow{2}{*}{\textbf{Hypothesis A}} & \multirow{2}{*}{\textbf{Hypothesis B}}\\ 
        &  \textbf{information} & & & \\\hline
        \multirow{2}{*}{(A)}   &  \multirow{2}{*}{WER = }& et on découvre les spectateurs & \underline{$\epsilon$} on découvre les spectateurs & et on découvre les \underline{spectacles} \\
        & & {\small \textcolor{gray}{and they discover the spectators}} & {\small \textcolor{gray}{\underline{$\epsilon$} they discover the spectators}} & {\small \textcolor{gray}{and they discover the show}} \\
        \multirow{2}{*}{(A)}     &  \multirow{2}{*}{CER $>$} & sur la vie politique & \underline{$\epsilon$} la vie politique & \underline{c'} la vie politique \\ 
        & & {\small \textcolor{gray}{on the political life}} & {\small \textcolor{gray}{\underline{$\epsilon$} the political life}} & {\small \textcolor{gray}{t's the political life}} \\
        \multirow{2}{*}{(A)}     &  \multirow{2}{*}{SemDist $>>$} & c' est à paris & \underline{$\epsilon$} est à paris & c' est \underline{appau} \underline{$\epsilon$} \\ 
        & & {\small \textcolor{gray}{it's at paris}} & {\small \textcolor{gray}{\underline{$\epsilon$} is at paris}} & {\small \textcolor{gray}{it's atpau \underline{$\epsilon$}}} \\
        \multirow{2}{*}{(B)}     &  WER = ;   & encore du rock & \underline{corps} du rock & encore du \underline{rok} \\ 
        & SemDist $>$& {\small \textcolor{gray}{still rock}} & {\small \textcolor{gray}{body of rock}} & {\small \textcolor{gray}{still rok}} \\
        \multirow{2}{*}{(C)}     &  WER $\neq$   & où les passions sont si vives & \underline{$\epsilon$} les \underline{patients} sont si \underline{vive} & où les \underline{patients} sont si \underline{vifs} \\ 
        & BERTscore & {\small \textcolor{gray}{where passions are so vivid}} & {\small \textcolor{gray}{\underline{$\epsilon$} the patients are so vivid}} & {\small \textcolor{gray}{where the patients are so lively}} \\\hline
        \end{tabular}
    }
    \label{tab:ex_hyp}
\end{table*}

While text transcriptions are derived from speech, we also consider a {\bf Phoneme Error Rate (PhonER)}, which involves computing the Levenshtein distance between reference and hypothesis sequences of phonemes obtained using a text-to-phoneme converter\footnote{\url{https://github.com/Remiphilius/PoemesProfonds}}. 

\section{Side-by-side human evaluation protocol}%
\label{sec:human_eval}

This section describes the collection of the HATS corpus. The setup of the perceptual experiment is summarized in Section~\ref{sec:perceptual_experiment}, while the protocol for selecting automatic transcripts for human evaluation is described in Section~\ref{sec:stimuli_choice}.

\subsection{Perceptual experiment}
\label{sec:perceptual_experiment}

In our study, the side-by-side experiment involves presenting the subject with a manually transcribed reference to represent the speech, as well as two automatic transcripts, each produced by a different system. 
The automatic transcriptions always contained errors with respect to the reference. 
Each triplet comprised of a reference and two hypotheses is called a stimulus to which participants react in choosing their preferred hypothesis. In the following, \textit{stimuli} refers to the different triplets to which each participant was confronted. 

The experiment was made available online which allowed for participants to realize the task remotely and at their preferred time. 
They used a mouse to choose their preferred hypothesis according to the reference. The study utilized a minimal instruction protocol (See Figure~\ref{fig:perceptualexp}), which allowed participants to self-determine the criteria that were important in determining the quality of a transcript. Figure~\ref{fig:perceptualexp} illustrates the visual display presented to the subjects during the study.
The reference was in written form only, in order to allow a comparison of ASR-oriented metrics and human perception within the same context~\cite{vasilescu2012cross}.

To avoid possible biases, the stimuli were presented in a random order, both for the order of the triplet, and for the order of the two hypotheses (the same hypothesis can be A or B).

For this study, 143 online participants volunteered. 
Before starting with the evaluation, they filled out a questionnaire helping to assess their age, spoken languages, and level of education. All participants are fluent in French and have an average age of 34 years with a standard deviation of 13.5 years. In Figure~\ref{fig:spk_characterization} and Figure~\ref{fig:spk_education}, we can see the distribution of number of spoken languages and education level for our studied population. Each participant evaluated 50 triplets of transcripts in random order, for a total time of about 15 minutes per participant.

\subsection{Protocol for stimuli selection}
\label{sec:stimuli_choice}


The transcription triplets coming from the REPERE test corpus were not selected randomly. In this study, we attach great importance to the selection of stimuli 
and we decided to study human behavior and metrics in complex situation, i.e. where humans have difficulties to choose the best transcription.
In this context, the aim was to maximize the diversity of choices to be made: subjects had to choose among errors made by different systems (since it is unlikely that different systems produce identical errors). Also, it would be interesting to study the cases where the choice is easy for the automatic metrics, as well where ambiguous scores are obtained, or where two metrics disagree to determine which one of the two hypotheses is best.

Therefore, the following three criteria had to be respected: (1) both hypotheses must be different from each other and have at least one character that differs from the reference, (2) hypotheses from every system were contrasted with hypotheses from every other system, and (3) hypotheses pair selection was based on metric scores. 
The selection criteria (3) based on the metrics can be divided into three different categories: (A) each metric was compared to itself presenting either the same, a slightly different or a highly different score between the two hypotheses, (B) in both hypotheses the WER or CER were equal but WER or CER, EmbER, SemDist, BERTScore were different, (C) metrics were contrasted with opposing predictions of the better hypothesis (e.g. WER\textsubscript{(hypA)}~\textgreater~WER\textsubscript{(hypB)} but CER\textsubscript{(hypA)} \textless~CER\textsubscript{(hypB)}). 

Table~\ref{tab:ex_hyp} illustrates how hypotheses were matched with concrete examples. 



\section{HATS data set}
\label{sec:HATS_set}

\subsection{Corpus description}
\label{sec:HATS}

The HATS data set includes 1,000 references, each with two different erroneous hypotheses generated by different ASR systems. The preferred choice of 143 human evaluators for each 50 reference-hypotheses triplets is recorded in this data set, resulting in a total of 7,150 annotations. Each triplet is evaluated by at least 7 participants.

To assess the level of agreement between raters, we calculate Fleiss' Kappa, which yields a value of 0.46. In 82\% of the triplet cases, the agreement (See Equation \ref{eq:agreement}), is found to be at least 71.4\%. Furthermore, in 60\% of the triplet cases, the agreement reaches a minimum of 85.7\%. This shows that the task is difficult, but that humans are still capable of determining a hypothesis as the best.

\begin{table*}[h!]
\centering
\begin{tabular}{lrrrrrr}
\hline
    & \multicolumn{2}{c}{\textbf{100~\%}} & \multicolumn{2}{c}{\textbf{70~\%}} & \multicolumn{2}{c}{\textbf{Total}} \\
    & Agree & Equal & Agree & Equal & Agree & Equal \\ \hline

\textbf{\textit{Lexical}} \\
Word Error Rate                & 63~\% & \textcolor{gray}{23~\%}   & 53~\% & \textcolor{gray}{28~\%}   & 49~\% & \textcolor{gray}{28~\%}         \\
Character Error Rate             & 77~\% & \textcolor{gray}{17~\%}    & 64~\% & \textcolor{gray}{21~\%}   & 60~\% & \textcolor{gray}{22~\%}         \\
\textbf{\textit{Semantic}} \\
Embedding Error Rate             & 73~\% & \textcolor{gray}{12~\%}    & 62~\% & \textcolor{gray}{16~\%}   & 57~\% & \textcolor{gray}{17~\%}         \\
BERTScore BERT-base multilingual & 84~\% & \textcolor{gray}{\hphantom{1}0~\%}     & 75~\% & \textcolor{gray}{\hphantom{1}1~\%}    & 70~\% & \textcolor{gray}{\hphantom{1}1~\%}          \\
BERTScore CamemBERT-base         & 81~\% & \textcolor{gray}{\hphantom{1}0~\%}     & 72~\% & \textcolor{gray}{\hphantom{1}0~\%}    & 68~\% & \textcolor{gray}{\hphantom{1}0~\%}          \\
BERTScore CamemBERT-large        & 80~\% & \textcolor{gray}{\hphantom{1}0~\%}     & 68~\% & \textcolor{gray}{\hphantom{1}0~\%}    & 65~\% & \textcolor{gray}{\hphantom{1}0~\%}          \\
SemDist CamemBERT-base           & 86~\% & \textcolor{gray}{\hphantom{1}0~\%}     & 74~\% & \textcolor{gray}{\hphantom{1}0~\%}    & 70~\% & \textcolor{gray}{\hphantom{1}0~\%}          \\
SemDist CamemBERT-large          & 80~\% & \textcolor{gray}{\hphantom{1}0~\%}     & 71~\% & \textcolor{gray}{\hphantom{1}0~\%}    & 67~\% & \textcolor{gray}{\hphantom{1}0~\%}          \\
SemDist Sentence CamemBERT-base  & 86~\% & \textcolor{gray}{\hphantom{1}0~\%}     & 75~\% & \textcolor{gray}{\hphantom{1}0~\%}    & 71~\% & \textcolor{gray}{\hphantom{1}0~\%}          \\
SemDist Sentence CamemBERT-large & 90~\% & \textcolor{gray}{\hphantom{1}0~\%}     & 78~\% & \textcolor{gray}{\hphantom{1}0~\%}    & 73~\% & \textcolor{gray}{\hphantom{1}0~\%}          \\
SemDist Sentence multilingual    & 76~\% & \textcolor{gray}{\hphantom{1}0~\%}     & 66~\% & \textcolor{gray}{\hphantom{1}0~\%}    & 62~\% & \textcolor{gray}{\hphantom{1}0~\%}          \\
SemDist FlauBERT-base            & 65~\% & \textcolor{gray}{\hphantom{1}0~\%}     & 62~\% & \textcolor{gray}{\hphantom{1}0~\%}    & 59~\% & \textcolor{gray}{\hphantom{1}0~\%}          \\

\textbf{\textit{Phonetic}} \\
Phoneme Error Rate            & 80~\% & \textcolor{gray}{14~\%}     & 69~\% & \textcolor{gray}{16~\%}    & 64~\% & \textcolor{gray}{17~\%}          \\ \hline
\end{tabular}
\caption{Performance of each metric according to their human agreement. \textbf{Full} means that no filter on agreement were applied on data set. Column "Equal" indicates the percentage of times the metric gave the same score to both hypotheses.}
\label{tab:cert_scores}
\end{table*}

\subsection{Methodology to evaluate metrics}
\label{sec:methods_metric_eval}

Our method for evaluating metrics involved calculating the proportion of instances where both human annotators and the metric selected the same hypothesis as the best option.
Subjects were not allowed to determine that the two hypotheses were equal. However, it is certain that there are cases where one hypothesis cannot be chosen and the subject chooses randomly. Since the number of annotators for each triplet is 90\% of the time odd, there will still be a winning hypothesis due to chance. One strategy to overcome this problem may be to take into account only the cases where there is a consensus.
In this study, we calculate a human agreement that corresponds to a percentage indicating consensus. This is calculated according to the following formula: 
\begin{equation}
\frac{\max(\mathit{A}, \mathit{B})}{\mathit{A} + \mathit{B}}
\label{eq:agreement}
\end{equation} 
where \textit{A} is the number of humans who select one hypothesis, and \textit{B} is the number of humans who select the other one. When agreement is weak, agreement is close to 50\%, and if all humans agree on the same hypothesis, agreement is 100\%. 
A filter can be applied on the data set according to three values of agreement: \textbf{100\%} (keep only triplets where all subjects agree), \textbf{70\%}, or \textbf{0\%} (no filter applied); which corresponds to 371, 819 and 1000 utterances respectively. The 70\% threshold was chosen in order to have consistent annotator agreement even if not all participants answer in the same way~\cite{nowak2010reliable}. 
Taking the predictions of the metrics as a starting point, we calculate the number of times that humans chose the best hypothesis based on the evaluated metric. 



\section{Evaluation of ASR metrics from human perspective}
\label{sec:metric_evaluation}

Table~\ref{tab:cert_scores} presents the results obtained by each metric according to the number of times they agree with human perception. Without surprise, the higher is the human agreement, the higher are the metrics performances. Unlike the results of previous studies~\cite{kim2021evaluating}, our study found that CER aligns more closely with human perception than WER. This divergence might be attributed to the use of written text as a reference in our perceptual experiment, rather than audio, or to intrinsic linguistic variations between French and English (French orthography contains a high number of silent letters compared to English).

It is interesting to note that at phoneme level, PER performs well, better than WER and CER despite the fact that humans have made their choices based on text alone. It shows that humans seem to consider how sentences sound even while reading. 
This is especially true if sentences are contrasted with a reference. 

Although hypotheses selected based on BERTScore using BERT-base-multilingual perform 8\% better than those chosen with SemDist Sentence multilingual, it would be premature to conclude that the BERTScore strategy is superior for evaluating the quality of transcripts as both metrics use different embeddings. When comparing these metrics with the same embeddings, SemDist outperforms BERTScore using CamemBERT-base embeddings while SemDist has a similar performance with BERTScore using CamemBERT-large. This suggests that some embeddings are more optimized for specific metrics.

On the 70\% and 0\% agreement level, WER have performances close to a random choice. This is due to the fact that in our data set, many cases present hypotheses with the same WER, and equal predictions are considered as a failure of the metric since humans are able to faithfully select one hypothesis.
Furthermore, we can observe that SemDist using FlauBERT-base embeddings performs worse than CER. This highlights the importance of carefully selecting embeddings and evaluating them on data sets like HATS before drawing conclusions about system performances at a semantic level. Based on our human-oriented data set, the best metric is SemDist using Sentence CamemBERT-large, which can be explained by the fact that this metric is based on embeddings specifically trained to maximize the similarity between sentences with similar meanings. It is worth noting that a large amount of annotated data is necessary to use these embedding-based metrics.


\section{Conclusion and Perspectives}
\label{sec:conclusion}



In this study, automatic evaluation metrics applied to transcriptions coming from different ASR systems were compared to human evaluation of different erroneous hypotheses according to one written reference. Our results show that SemDist with Sentence-BERT evaluates transcripts in a way that seems acceptable for human raters. If Sentence-BERT is not a possible option, BERTScore seems to be the second best option. This metric is more stable than SemDist on BERT embeddings. Nevertheless, if possible, metrics should be evaluated through data sets comprising also human annotations such as HATS. 


Although these new evaluation methods are interesting in the context of ASR, the advantage of WER and CER metrics lies in their computational low-cost and interpretability of the score. Therefore, the next step could be to develop metrics that correlate with human perception while remaining interpretable.

As future work, an additional study could be conducted by replicating the current experiment using an audio reference instead of a textual reference, so that subjects do not have character information. This approach would enable us to examine any variations and if CER is still considered as better than the WER in a multimodal setting.


\section{Limitations}
\label{sec:limitations}


The HATS data set is not necessarily representative of all kind of errors nor the most common because errors were selected applying strict criteria. In order to evaluate the representativeness of this data set, additional analyses with respect to the kind of errors that occur in each system's transcriptions have to be carried out. 

Furthermore, conclusions drawn from this data set may be specific to the French language and may not generalize to other languages. Adding and comparing similar data sets in other languages would help to better understand the performance of metrics and human evaluations across different languages.

\section{Ethics Statement}

The aim of this paper is to propose a new method for evaluating speech-to-text systems that better aligns with human perception. However, the inherent subjectivity of transcription quality means that if we optimize systems to correlate only with the perception of the studied population, it could be inequitable if this perception does not generalize to the rest of the population.


\section{Acknowledgements}
This work was supported by the DIETS project financed by the Agence Nationale de la Recherche (ANR) under contract ANR-20-CE23-0005. It was granted access to the HPC resources of IDRIS under the allocation 2021-A0111012991 made by GENCI.

\bibliographystyle{IEEEtran}
\bibliography{mybib}

\end{document}